\documentclass{jpsj3}
\usepackage{txfonts}
\usepackage{color}
\title{Detection of cheating by decimation algorithm}

\author{Shogo Yamanaka$^1$\thanks{yamanaka.shogo.74u@st.kyoto-u.ac.jp}, Masayuki Ohzeki$^2$, and Aur\'elien Decelle$^3$}
\inst{$^1$Department of Engineering, Kyoto University, 36-1 Yoshida Hon-machi, Sakyo-ku, Kyoto, 606-8501, Japan \\
$^2$Department of Systems Science, Graduate School of Informatics, Kyoto University, 36-1 Yoshida Hon-machi, Sakyo-ku, Kyoto, 606-8501, Japan \\
$^3$Dipartimento di Fisica, Universit\`a La Sapienza, Piazzale Aldo Moro 5, I-00185 Roma, Italy} %\\

\abst{We expand the item response theory to study the case of ``cheating students" for a set of exams, trying to detect them by applying a greedy algorithm of inference. This extended model is closely related to the Boltzmann machine learning. 
In this paper we aim to infer the correct biases and interactions of our model by considering a relatively small number of sets of training data.
Nevertheless, the greedy algorithm that we employed in the present study exhibits good performance with a few number of training data. The key point is the sparseness of the interactions in our problem in the context of the Boltzmann machine learning: the existence of cheating students is expected to be very rare (possibly even in real world).
We compare a standard approach to infer the sparse interactions in the Boltzmann machine learning to our  greedy algorithm and we find the latter to be superior in several aspects.
}

\begin{document}

\maketitle

\section{Introduction}
The Boltzmann machine learning (or equivalently the inverse Ising problem), which is one of the methods in statistical machine learning theory\cite{Ackley1985}, is a useful tool to describe data issued from strongly correlated systems.
Recently, many developments have been done as well on the algorithmic part by elaborating efficient methods (for instance in computer science \cite{raskutti2008model}, or as well in physics\cite{Sessak2009,nguyen2012bethe,Ricci2012,Aurelien2014}), but also on the experimental part where many data became available to study as in neural network\cite{cocco2009neuronal} or in  biology\cite{morcos2011direct}. This growing number of data needs an efficient inference process to understand quantitatively the nature of the studied system and to describe correctly the observed complex behavior.
The generative model assumed in the Boltzmann machine learning takes the form of a probability density defined by using the Hamiltonian of the Ising model. This model contains both a bias on each variable as well as pairwise interactions between the different variables.
However, at the stage of learning, the Boltzmann machine learning demands a relatively large number of training data. Moreover, the computation of the likelihood of the model has a very high computational cost (it has an exponential complexity with the system size).
One of the simplest way to mitigate the latter difficulty in the Boltzmann machine learning is to use the pseudo likelihood estimation \cite{Besag1975,Ekeberg2013}, which asymptotically (large number of training data) coincides with the maximum of the likelihood of the problem.
The other way is to construct a good approximation which infers well the biases and interactions even when the number of samples in the training data is small \cite{Sessak2009,Cocco2011,Cocco2012,Ricci2012,Yasuda2013,Raymond2013,Ohzeki2013}.
In addition, we may sometimes have a prior knowledge on the structure of the data which we could take into account in the inference process. Therefore, taken into consideration this prior knowledge, it enables us to circumvent the week point of the pseudo likelihood estimation, which demands a vast number of the training data to make the inference precise. 
If one can expect that the underlying structure of interactions described by the data is sparse, as for the data we deal with in the present study, a greedy approach can give good inference in conjunction with the simple pseudo likelihood estimation \cite{Aurelien2014}.
The algorithm allows then to significantly reduces the number of training data to achieve an efficient learning where most of the interactions in the generative model have been put to zero.

In the present study, we apply the greedy method to an extended model of the generative model based on the item response theory \cite{Baker2004}.
The item response theory is a probabilistic technique used to estimate the ability of a group of examinees to succeed or fail to a group of tests of various difficulties.
The method is usually employed to assess the validity and efficiency of a kind of certification tests and keep their quality.
The applicability of the item response theory does not restrict itself to the specialized type of the tests.
Any kind of examinations, which are conducted in universities, are in the range of the item response theory.
Although the usual setting on the item response theory does not assume the existence of ``cheating students", we demonstrate here the possibility to detect them by applying inference methods on a sample of answer sheets corresponding to a series of tests.

The present paper consists of the following sections.
The second section formulates the item response theory with cheating students in context of the Boltzmann machine learning. 
We give a brief introduction of our technique, namely the decimation algorithm, in the third section.
In the next section, the numerical experiments demonstrate good performances of our algorithm to detect the existence of the cheating students.  It is also efficient in inferring the ability of the examinees and the difficulty of the problems they solved.

\section{Item response theory and its extension}
The item response theory is proposed as a method to estimate the ability of each examinee to succeed to a series of tests of various difficulties.
It is often employed in various qualifying examinations as well as in the context of sociology and psychology to specify the ability of the examinees.
The item response theory introduces a probability distribution to express how likely it is that the examinees resolve a series of problems according to their individual ability and the difficulty of the problems.
In the theory, we use a logistic function formed as a template to express the relationship between the answers from the examinees with their ability to the problems and their difficulty.
In our formulation, the independence between the problems and between the examinees is assumed.
We define the ability of the $i$th examinee ($i=1,2,\cdots,I$) as $\theta_i$ and the difficulty of the $j$th problem ($j=1,2,\cdots,J$) as $d_j$.
The probability that all of the examinees give the answers ${\bf x}$ is expressed as
\begin{equation}
P({\bf x}|\boldsymbol{\theta}, {\bf d}) = \frac{1}{Z(\boldsymbol{\theta}, {\bf d})}\prod_{j=1}^J \exp\left\{ \sum_{i=1}^I (\theta_i - d_j)x_{ij} \right\},
\label{1PL}
\end{equation}
where $x_{ij}$ is the result of the answer by the $i$th examinee on the $j$th problem and we define $x_{ij} = 1$ if the the answer is correct and $x_{ij} = -1$ otherwise.
We here define the constant for the normalization as $Z(\boldsymbol{\theta}, {\bf d})$.
This model is called the one-parameter logistic (1PL) model in the context of the item response theory.
In general, extended version of the model are proposed to get a better description of the probability on the results of the examinees.
For simplicity, we use the 1PL model here to infer the ability of the examinees and the difficulty of the problems they solved.

The function (\ref{1PL}) can also be seen as the likelihood of the inferred parameters.
The standard procedure to infer the parameters is to consider the maximization of the log-likelihood function $\sum_{s=1}^N \log P({\bf r}^{(s)}|\boldsymbol{\theta}, {\bf d})$, where we insert ${\bf r}^{(s)}$ into the argument and $N$ denotes the number of samples in the data set.
In the ordinary setting of the item response theory, we assume that the examinees are independent.
We expand the ordinary setting of the item response theory to the case where some ``cheating students" exist by considering the existence of positive-valued interactions between several particular pairs as
\begin{equation}
P({\bf x}|\boldsymbol{\theta}, {\bf d},{\bf w}) = \frac{1}{Z(\boldsymbol{\theta}, {\bf d}, {\bf w})} \prod_{j=1}^J\exp \left\{\sum_{i=1}^I (\theta_i - d_j)x_{ij} + \sum_{i=1}^I\sum_{ k\in \partial (i)} w_{ik} x_{ij}x_{kj} \right\},\label{eq:1}
\end{equation}
where $w_{ik}$ is a coefficient of ``cheating pair" to express the correlation between a pair of students $ik$.
When a cooperative pair cheats, $w_{ik}$ takes a positive value. Otherwise $w_{ik} = 0$.
We assume $w_{ik} = w_{ki}$, namely sharing the information to cheat on tests.
Here we define the normalization factor as
\begin{equation}
Z(\boldsymbol{\theta}, {\bf d}, {\bf w}) \equiv   \sum_{{\bf x}_j} \prod_{j=1}^J \exp \left( -E({\bf x}_j | {\bf \theta}, d_j, {\bf w} )\right) ,
\end{equation}
where ${\bf x}_j = (x_{1j},x_{2j},\cdots,x_{Ij})$ and 
\begin{equation}
E({\bf x}_j | \boldsymbol{\theta}, d_j, {\bf w}) \equiv -\sum_{i=1}^{I} (\theta_i - d_j) x_{i j} - \sum_{i=1}^I \sum_{k \in \partial (i) } w_{i k} x_{i j} x_{k j}.
\end{equation}
This is the same function as the standard form of the Hamiltonian of the random-field and random-bond Ising model.
By introducing interactions between the problems, we can also deal with the case where several problems are constructed by using a series of related questions \cite{Yasuda2012}.
The introduction of the interactions $w_{ik}$ is reasonable.
In order to the examinee $i$ to correctly answer to the problem $j$, its ability $\theta_i$ must exceed the difficulty of the $j$th problem $d_j$.
However, if he shares information with the $k$th examinee, meaning having $w_{ik}>0$, the extra term $w_{ik}x_{kj}$ will help the $i$th examinee to succeed to the problem $j$. 
The summation over $\partial (i)$ means the adjacent examinees of the $i$th examinee which in practical case could be tuned to take into account how the tests are done.
For instance, a reasonable assumption in the case where the tests are performed in an exam room on desks would be to locate the examinees at the sites of a square lattice.
Another example could be a situation where we demand to the examinees to hand in some reports after a few days. In that case, all the examinees can be in contact with each other leading to consider that everybody is ``connected'' to everybody (as in infinite range models).

\subsection{Parameter estimations}
The goal of the original formulation of the item response theory is to estimate the ability of the examinees $\boldsymbol{\theta}$ and the difficulty of the problems ${\bf d}$ from the given score data ${\bf r}$. 
In our model, the aim straightforwardly corresponds to the Boltzmann machine learning.
Detecting  the existence of cheaters is interpreted as inference of the cheating coefficient ${\bf w}$ from the given data ${\bf r}$ in the formalism of the Boltzmann machine learning. 

Let us tackle the inference problem on $\boldsymbol{\theta}$, ${\bf d}$, and ${\bf w}$ from the data ${\bf r}$ generated from Eq. (\ref{eq:1}). Notice that in Eq. (\ref{eq:1}) $P({\bf x}|\boldsymbol{\theta},{\bf d},{\bf w}) = P({\bf x}|\boldsymbol{\theta}+ \boldsymbol{\eta},{\bf d} + \boldsymbol{\eta},{\bf w})$ for the arbitrary real values $\boldsymbol{\eta}$.
Thus we cannot identify $\boldsymbol{\theta}$ and ${\bf d}$ uniquely.
In order to avoid this arbitrariness, we are going to follow an idea of a previous study \cite{Yasuda2012}, where it is assumed that the $d_j$ are independent random variables generated from the Gaussian distribution ${\cal N}(d_j | \mu, \sigma^{2})$.
We take the maximizers of the following joint probability as the most likely estimations of $\boldsymbol{\theta}$,$ {\bf d}$ and $ {\bf w}$.
\begin{equation}
P_{\text{joint}}(\mathbf{r},\mathbf{d}|\boldsymbol{\theta}, \mathbf{w}) \equiv P(\mathbf{r} | \boldsymbol{\theta},\mathbf{d},\mathbf{w})
 \prod_{j=1}^{I} {\cal N}(d_j | \mu , \sigma^2 ),
\end{equation}
It is then convenient to maximize its logarithm, instead of the joint probability, 
\begin{eqnarray}
{\cal L} (\boldsymbol{\theta} ,{\bf d}, {\bf w})&\equiv& \ln
 P_{\text{joint}} ({\bf r},{\bf d}|\boldsymbol{\theta},{\bf w}) \notag\\
&= &\sum_{j=1}^{J}\left(\ln P({\bf r}_j|{\bf \theta},d_j,{\bf w})- \frac{(d_j - \mu)^2}{2\sigma^2}\right). \label{LogLike} 
\end{eqnarray}
where $\ln P({\bf r}_j|{\bf \theta},d_j,{\bf w}) = - E({\bf r}_j|{\bf \theta},d_j,{\bf w}) - \log Z_j({\bf \theta},d_j,{\bf w})$.
It is however difficult in general to maximize the log-likelihood function (\ref{LogLike}). 
Let us introduce the pseudo log-likelihood function by approximating the first term in Eq. (\ref{LogLike}) as
\begin{equation}
\ln P(\mathbf{r}_j | \boldsymbol{\theta}, d_j,{\bf w}) \approx
 \sum_{i=1}^{I} \ln P(r_{ij}| \boldsymbol{\theta},d_j,{\bf w},{\bf r}_{j\setminus i}), 
\end{equation}
where
\begin{eqnarray}
P(r_{ij}| \boldsymbol{\theta},d_j,{\bf w},{\bf r}_{j\setminus i}) &\equiv&
 \frac{\exp \left(-E(\mathbf{r}_j|\boldsymbol{\theta}, d_j, {\bf w}   )\right)}{\sum_{\{\mathbf{r}_j\}}
 \exp\left(-E({\bf r}_j| \boldsymbol{\theta}, d_j, {\bf w}    )\right)} \notag\\
&=& \frac{\exp\left\{ \left(\theta_i -d_j + \sum_{k \in \partial(i)} w_{ik} r_{kj}\right)r_{ij}\right\}}{2\cosh \left\{\left(\theta_i -d_j + \sum_{k \in \partial(i)} w_{ik} r_{kj}\right)r_{ij}\right\}} 
\end{eqnarray}
and ${\bf r}_{j\setminus i} := {\bf r}_j\setminus \{r_{ij}\}$.
Then Eq. (\ref{LogLike}) is reduced to 
\begin{eqnarray}
{\cal L} (\boldsymbol{\theta},{\bf d},\mathbf{w}) &\approx& \sum_{i=1}^I \sum_{j=1}^{J} (\theta_i - d_j)r_{ij} + \sum_{i=1}^I
 \sum_{j=1}^{J} \sum_{k \in \partial(i)}^{I} w_{ik}r_{ij} r_{kj} \notag\\
&&\quad -\sum_{i=1}^{I} \sum_{j=1}^J \ln 2\cosh \left(\theta_i -d_j +\sum_{k\in
 \partial{(i)}} w_{ik} r_{kj} \right) -\sum_{j=1}^{J} \frac{(d_j -\mu)^2}{2\sigma^2} \notag\\
&\equiv& {\cal PL}(\boldsymbol{\theta}, {\bf d}, {\bf w}). \label{eq:2}
\end{eqnarray}
We call the function in Eq. (\ref{eq:2}), ${\cal PL}$, the pseudo log-likelihood function.
The method to infer the parameters by maximizing the pseudo log-likelihood function is ``Pseudo Likelihood Maximization'' (PLM)\cite{Besag1975}.
We may employ the steepest descent method to maximize it due to strict convexity of the pseudo log-likelihood function.
We give the gradient of ${\cal PL}(\boldsymbol{\theta},{\bf d},{\bf w})$ as
\begin{eqnarray}
\frac{\partial {\cal PL} (\boldsymbol{\theta},{\bf d},{\bf w})}{\partial \theta_i} &=& \sum_{j=1}^J r_{ij} - \sum_{j=1}^J {\cal A}_{ij} \label{grad_theta}\\
\frac{\partial {\cal PL} (\boldsymbol{\theta},{\bf d},{\bf w})}{\partial d_j} &=&-\sum_{i=1}^{I} r_{ij} + \sum_{i=1}^{I} {\cal A}_{ij} - \frac{d_j - \mu}{\sigma^2} \label{grad_d}\\
\frac{\partial {\cal PL} (\boldsymbol{\theta},{\bf d},{\bf w})}{\partial w_{ik}} &=& 2\sum_{j=1}^{J} r_{ij}r_{kj} - \sum_{j=1}^{J} r_{kj}{\cal A}_{ij} - \sum_{j=1}^J r_{ij}{\cal A}_{kj},  \label{grad_w}
\end{eqnarray}
where
\begin{equation}
{\cal A}_{ij} \equiv \tanh \left(\theta_i - d_j + \sum_{k\in \partial (i)} w_{kj} r_{ij} \right).
\end{equation}
We use a gradient descent method by using Eqs. (\ref{grad_theta}-\ref{grad_w}) to update the parameters  $\boldsymbol{\theta}$, ${\bf d}$ and ${\bf w}$ until we reach the maximum of the pseudo likelihood function.
The obtained values are the results of the PLM estimation.
In the present formulation, we assume that each problem is independent from each other.
This means that the set of tests can be interpreted as an independent set of data to infer the parameters, namely $J=N$.
Therefore, the precision on the estimate of the parameters is increased when the number of tests $J$ is increased as well.

\section{Utilization of Sparseness}
\subsection{Sparseness}
When using the item response theory to detect the existence of cheating pairs between examinees in the given data, we may assume that most of the cheating coefficients are zeros.
It is sparsity. A common method to deal with inferring the parameters of sparse system is to use the $L_1$ norm as a prior distribution on the parameters\cite{Bishop2006}.
This  prior distribution for ${\bf w}$ is then multiplied by the likelihood to obtain the posterior distribution:
\begin{equation}
P_{\text{joint}}(\mathbf{r},\mathbf{d},\mathbf{w}|\boldsymbol{\theta}) \equiv P_{\text{joint}}(\mathbf{r},\mathbf{d}|\boldsymbol{\theta}, \mathbf{w})P({\bf w}),
\end{equation}
where $P(\mathbf{w})$ is set to be $P(\mathbf{w}) \propto \exp\left(-\lambda|{\bf w}|\right)$.
Instead of the maximization of $P_{\text{joint}}(\mathbf{r},\mathbf{d}|\boldsymbol{\theta}, \mathbf{w})$, we find the maximizer of $P_{\text{joint}}(\mathbf{r},\mathbf{d},\mathbf{w}|\boldsymbol{\theta})$.
In the rest of the article we denote this procedure as PLM+$L_1$ estimation.
One crucial remaining problem on the PLM+$L_1$ estimation is the arbitrariness of the parameter $\lambda$.
In order to avoid this arbitrariness, we will employ the decimation algorithm. This puts the parameters of the model (here $w_{ik}$) to zero iteratively. Each step of the algorithm consists in  maximizing the pseudo likelihood and then to put to zero (or decimate) a fraction of the smallest parameters. This method was proposed in  a previous study \cite{Aurelien2014} and we will refer to it as PLM+decimation in the rest of our work.
This method does not suffer from the arbitrariness of some parameter such as the one on the regularization $L_1$ norm. In addition, it shows outstanding performance and beat PLM+$L_1$ in many situations.
In the present study, we apply the PLM+decimation to our model in order to infer the rare existence of cheating on tests.
We assess the performance of PLM+decimation in detail by evaluating the inference of the ability of the examinee and the difficulty of the problem as well as the detection of cheating tests.

\subsection{Decimation algorithm}
The pseudo log-likelihood function should strongly depend on whether $w_{ik}$ takes a non-zero value or not.
We thus divide the group of $w_{ik}$ into two subgroups.
One is $w_{ik}$ with non-zero values as ${\bf W}_{1}$ and the other is $w_{ik}$ with zero values as ${\bf W}_{0}$.
In the PLM+decimation algorithm, a fraction of the system is put to zero at each step. The parameters $w_{ik}$ with a small value are interpreted of potentially being absent and can be assigned to the group of inferred parameters that are zero:  ${\bf W}^*_0$. Here the asterisk indicates that we refer to the inferred parameters.
At the first stage of inference by PLM+decimation, we initially set ${\bf W}^*_0 = \emptyset$.
At each step of maximization of ${\cal PL}$, we assign a set of $w_{ik}$ to ${\bf W}^*_0$ by a fixed ratio of $\rho$. 
We call this procedure ``decimation''.
We have now to define a criterion telling us when to stop the decimation process.
In other words, we must find a way to stop the process as close as possible of the point where ${\bf W}^*_0 = {\bf W}_0$.
The decimation can be seen as forcing the parameters belonging to  ${\bf W}_0^*$ to be zero: $w_{ik}^* = 0$ if $w_{ik}^*\in {\bf W}_0^*$ during the remaining steps of the inference and when maximizing the pseudo log-likelihood.
Thus we find that the following inequality holds during PLM+decimation algorithm.
\begin{equation}
 {\cal PL}_{{\rm min}} \leq {\cal PL} \leq {\cal PL}_{\rm{ max}},
\end{equation}
Here ${\cal PL}_{{\rm min}}$ is the value of the pseudo log-likelihood function when all parameters (the $w_{ik}$) have been put to ${\bf W}^*_{0}$. 
On the other hand, ${\cal PL}_{{\rm max}}$ is the value of the pseudo log-likelihood when all the $w_{ik}$ are present (i.e. ${\bf W}^*_0 = \emptyset$) and when the pseudo log-likelihood has been maximized over those parameters. 
Note that, the maximum value of a likelihood function cannot decrease when increasing the number of parameters (keeping all the previous ones) of the generative model since this ``larger'' model will include the ``smaller'' one. Therefore justifying the above inequality.
At this time, we can expect that, if we decimate all the correct null parameters, $w_{ik} \in {\bf W}_0$, the pseudo likelihood should not change drastically since these parameters are useless in principle to describe the data and therefore the difference between ${\cal PL}$ and ${\cal PL}_{{\rm max}}$ is very small.
On the other hand, if we decimate a coupling such that $w_{ik} \in {\bf W}_1$ and assign it to ${\bf W}_0^*$, the pseudo likelihood should change a lot and thus the difference between ${\cal PL}$ and ${\cal PL}_{{\rm max}}$ can be large.
In other words, at the early stage of the inference process using the PLM+decimation algorithm, ${\cal PL}$ takes values close to ${\cal PL}_{{\rm max}}$.
As the step increases, ${\cal PL}$ suddenly switches and approaches ${\cal PL}_{{\rm min}}$.
This sudden change is a signal that we are decimating the wrong parameters and should be a point where we are the closer possible to have ${\bf W}_0^* = {\bf W}_0$.
It is convenient to define the more sharp quantity to represent the trigger as
\begin{equation}
{\cal PL}_{{\rm tilted}} = {\cal PL} - x {\cal PL}_{{\rm max}} - (1-x) {\cal PL}_{{\rm min}}.
\end{equation}
where $x$ is the ratio of decimated parameters. This function vanishes when $x=0$ or $x=1$ and therefore present a maximum which we took as the signal to stop the decimation process. Further details on this function can be found in the literature\cite{Aurelien2014}.

\section{Numerical experiments}
We first perform numerical experiments on detection of cheating students by use of the PLM+decimation algorithm.
We set the number of examinees as $I=30$ and we therefore have $I(I-1)/2 = 435$ potential pairs of cheaters.
To control the dilution of the true model, we will denote $p$ the ratio
of cheating pairs that we introduce. 
We set the coefficient $w_{ij} = 1$ with a probability $p$ and $w_{ij} = 0$ with $1-p$.
The ability of the examinee $\theta_i$ is generated from ${\cal N}(0,0.5)$ and the difficulty of the problems $d_j$ is also distributed following ${\cal N}(0,0.5)$.
We then generate the score data ${\bf r}$ according to Eq. (\ref{eq:1}) by use of the standard Markov-chain Monte-Carlo method.
By use of the obtained score data, we maximize the pseudo likelihood function following the procedure of the PLM+decimation.
In the standard PLM estimation, the gradient descent step of the algorithm is done until the algorithm converges toward a given value. 
In the decimation algorithm, each step of the algorithm consists in a thousand iterations of the gradient descent algorithm to infer a first guess of the parameters. 
Then, we decimate a fraction of the system before going back to the gradient descent step. 
We set the ratio of the decimation step to be $\rho = 0.05$. 
 
\subsection{Results by PLM+decimation}
\begin{figure}[h]
\begin{center}
\includegraphics[width = 7cm]{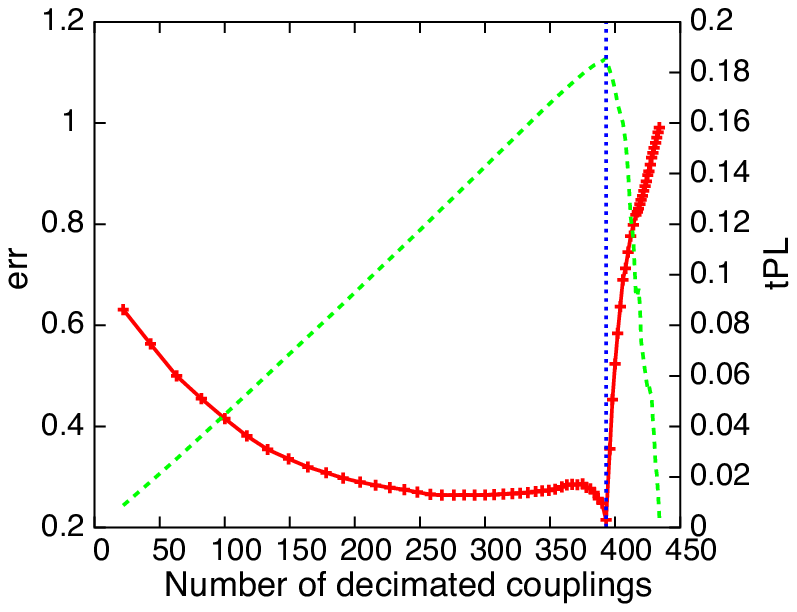}
\includegraphics[width = 7cm]{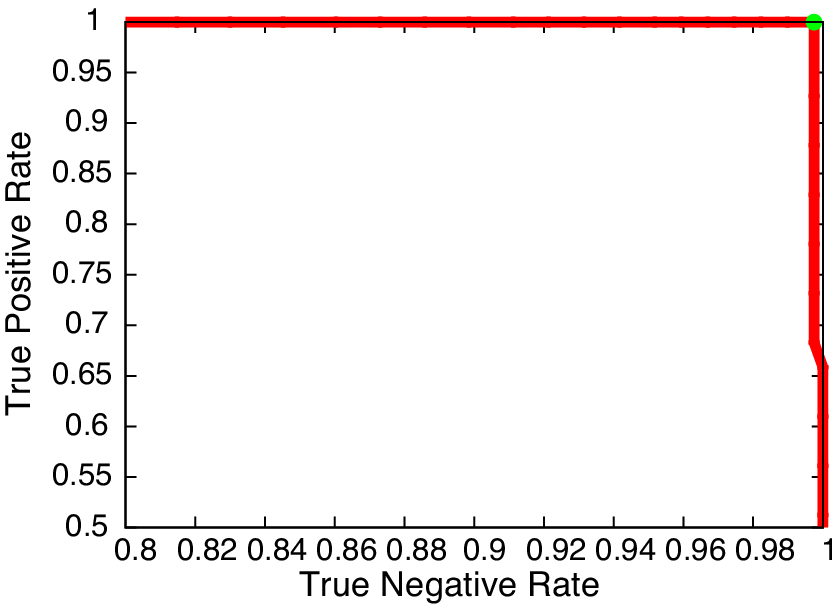}
\caption{ ${\cal PL}_{{\rm tilted}}$ and errors in PLM+decimation algorithm for the case with $p =0.1$ (left panel) and $J=1000$ and ROC curve (right panel).
We plot the value of ${\cal PL}_{{\rm tilted}}$ denoted by tPL in the green dashed curve.
In addition, the red solid curves describes the value of errors (${\rm err}_{w}$).
The minimum of errors coincides with the maximum point of ${\cal PL}_{{\rm tilted}}$.
We depict the maximum point of ${\cal PL}_{{\rm tilted}}$ by the green circle on the ROC curve.
The terminal point is located at the most right upper side in the ROC curve.} 
\label{fig:1}
\end{center}
\end{figure}

We show the results of PLM+decimation with $p = 0.1$ and $J=1000$ in Fig. $\ref{fig:1}$.
We measure ${\cal PL}_{{\rm tilted}}$ at each step and the error value is defined as
\begin{equation}
{\rm err}_{w} = \sqrt{\frac{\sum_{i < k} (w_{ik} -w_{ik}^*)^2}{\sum_{i<k} w_{ik}^2}},\label{err}
\end{equation}
where $w^*_{ik}$ is the inferred value generated by the numerical experiments.
As shown in Fig. $\ref{fig:1}$, the error value drastically changes when reaching the maximum of ${\cal PL}_{{\rm tilted}}$.
In addition, we put ROC curve in the right panel of Fig. \ref{fig:1}.
We define the true positive rate (TPR) as the ratio of the number of parameters $w_{ik} \in {{\bf W}_1}$ which are assigned to ${\bf W}_1^*$ and the number of parameters in ${\bf W}_1$. The true negative rate (TNR) is the same quantity defined for the null parameters. Therefore, on the ROC curve, the performance gets better and better when we approach the corner $(1,1)$ for the TPR and TNR.
The maximum of ${\cal PL}_{{\rm tilted}}$ is in general located at the most upper-right point in the ROC curve, which shows that both the TPR and TNR are close to unity, namely good estimations.

\subsection{Dependence on $p$}
\begin{figure}[h]
 \begin{center}
  \includegraphics[width=80mm]{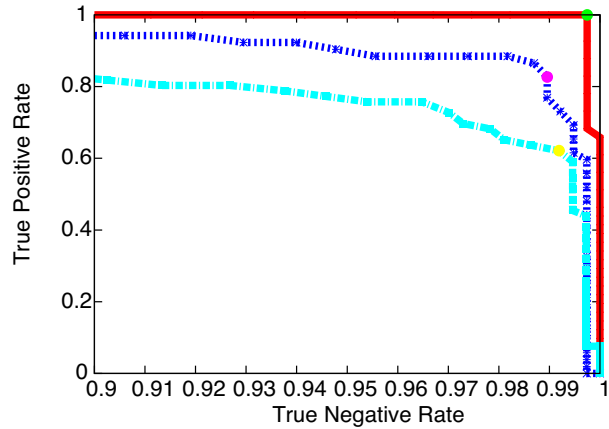}
 \end{center}
  \caption{ROC curves for the cases $p=0.1$, $0.125$ and $0.15$ from upper right to lower left when $J=1000$. 
As $p$ increases, TPR also decreases.}
  \label{fig:14}
\end{figure}

\begin{figure}[h]
 \begin{center}
  \includegraphics[width=80mm]{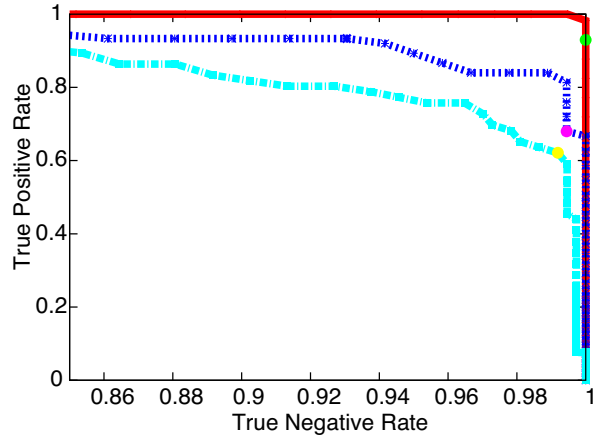}
 \end{center}
  \caption{ROC curves for the case with $J=1000$, $1600$ and $2000$  from lower left to upper right when $p= 0.15$.
Increase of $J$, PLM+decimation algorithm yields good estimations with larger TPR.}
  \label{fig:9}
\end{figure}

We investigate the dependence of the performance of PLM+decimation algorithm for various values of $p$.
We plot the ROC curves for the cases $p=0.1$, $0.125$ and $p=0.15$ in Fig. \ref{fig:14}. In all these cases, we keep a constant number of samples $J=1000$.
We can observe that, when $p$ increases, the TPR of the PLM+decimation gets a lower value.
We can therefore see that the PLM+decimation algorithm can work well on the sparse model where most of the parameters are zero. We remark also that, the density of the present cheaters increases (keeping the number of tests constant) the quality of the inference process gets reduced.
In Fig. \ref{fig:9} we illustrate the decimation algorithm at a fixed value $p=0.15$ and increasing the number of samples $J$. We clearly see that, for relatively large value of $p$, if we prepare more training data $J=1000$, $J=1600$ and $J=2000$, PLM+decimation algorithm leads to a better and better estimation with very few errors on the TPR at the end.

\subsection{Comparison to $L_1$ regularization}
In this section we compare the results obtained by PLM+decimation to  the PLM+$L_1$ method.
We recall that the $L_1$ regularization can be simply implemented by adding $\lambda \sum_{i<k}|w_{ik}|$ to Eq. (\ref{eq:2}), where $\lambda$ is the regularization coefficient. 
However, we have to mention that the performance of the $L_1$ regularization strongly depends on the value of $\lambda$. 
Indeed, by varying the value of $\lambda$, the results of the inference process change. 
For small values, we find that a lot of parameters are not put to zero whereas above a given threshold $\lambda_{\rm max}$, all parameters are pruned. 
Therefore, the good performance of PLM+$L_1$ should be taken with care since it is not possible in general to decide what would be the optimum value for $\lambda$. 

\begin{figure}[h]
 \begin{center}
  \includegraphics[width=80mm]{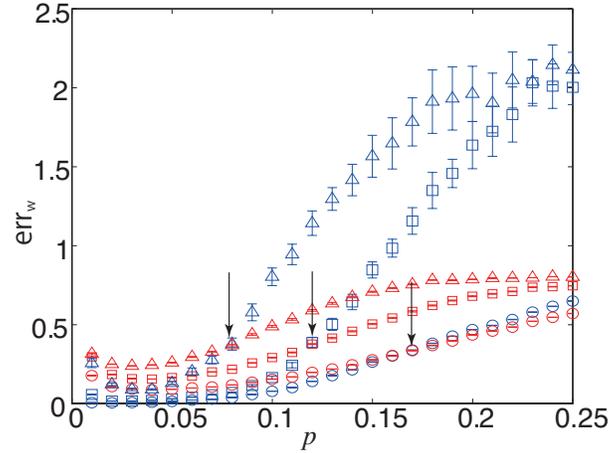}
 \end{center}
  \caption{Mean of errors in estimation of $w$ by PLM + decimation and PLM+$L_1$ estimation.
The triangles denote the results for $J=500$, the squares represent those for $J=1000$, and the circles stand for the case of $J=2000$.
The blue marks are by PLM + decimation algorithm, and the red ones are by PLM + $L_1$.
Three downward arrows represent the locations at which PLM+decimation outperforms PLM + $L_1$ in ${\rm err}_{w}$.
 We take the mean over $100$ samples for each method.
 We also put the error bars for each case.
 }
  \label{StatL1dec}
\end{figure}

\begin{figure}[h]
 \begin{center}
  \includegraphics[width=160mm]{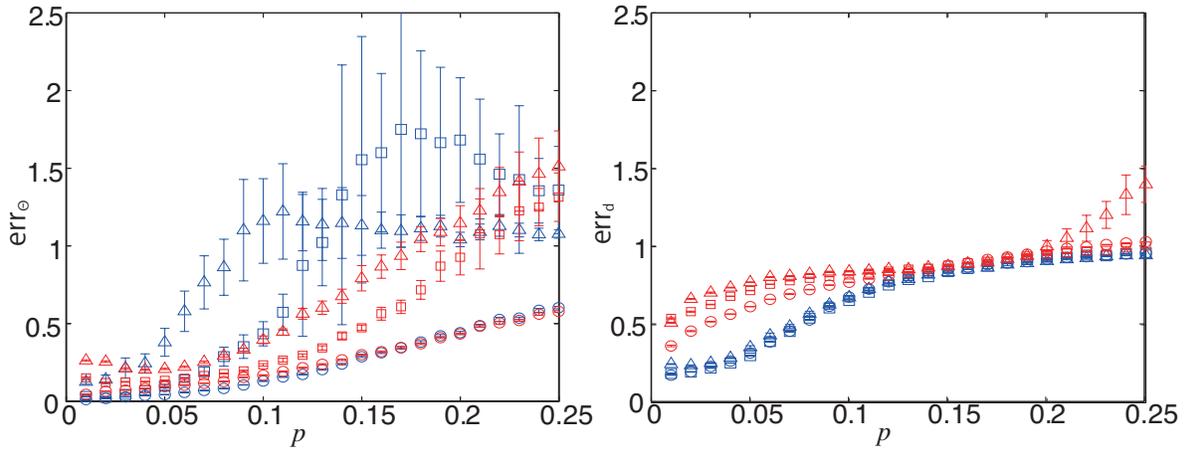}
 \end{center}
  \caption{Mean of errors in estimation of $\theta$ and $d$ by PLM + decimation and PLM+$L_1$ estimation.
  The same symbols are used as in Fig. \ref{StatL1dec}.
 We take the mean over $100$ samples for each method.
 We also put the error bars for each case.
 }
  \label{StatL1dec2}
\end{figure}

We show the comparative results obtained by PLM+decimation and PLM$+L_1$ in Fig. \ref{StatL1dec}. 
We run the numerical estimations in the $100$ samples for $I=30$ and $J=500,1000$ and $2000$, while tuning $p=0-0.25$.
The rate of the decimation is fixed to $\rho = 0.05$.
In the present experiment we choose the value of $\lambda$ such that the error on the cheating coefficient is minimized for each sample.
In this example, PLM+decimation outperforms clearly PLM$+L_1$ for the sparse cases in which most of the pairs do not cheat on the tests.
Increase of $J$ yields remarkable improvement on performance of PLM + decimation.
The location at which PLM+decimation outperforms PLM + $L_1$ in ${\rm err}_{w}$ moves to larger $p$ against increase of $J$.
Similarly to the case of ${\bf w}$ in Eq. (\ref{err}), we define the error values of $\boldsymbol{\theta}$ and ${\bf d}$ as ${\rm err}_{\theta}$ and ${\rm err}_d$ respectively:
\begin{eqnarray}
{\rm err}_{\theta} &=& \sqrt{\frac{\sum_{i=1}^I (\theta_i - \theta_i^*)^2}{\sum_{i=1}^I \theta_i^2}} \\
{\rm err}_{d} &=& \sqrt{\frac{\sum_{j=1}^J (d_j - d_j^*)^2}{\sum_{j=1}^J d_j^2}},
\end{eqnarray}
where $\theta_i^*$ and $d_j^*$ are the values inferred by the numerical experiment.
As shown in Fig. \ref{StatL1dec2}, PLM+decimation algorithm outperforms the PLM+$L_1$ estimation regarding the error values of $\boldsymbol{\theta}$ and ${\bf d}$.
The relatively large error bars are put on the errors of $\boldsymbol{\theta}$ due to lack of the regularization on this quantity.
The large number of $J$ reduces the uncertainty of the estimation even in $\boldsymbol{\theta}$.
We can therefore conclude that PLM+decimation algorithm is a good tool to detect the existence of cheating students and simultaneously infer the ability of the examinees and the difficulty of the problems in terms of the item response theory.
We emphasize that, although our model corresponds to the Ising model, namely the ordinary Boltzmann machine learning, the estimated quantities are not only the magnetic fields and the interactions. 
In our model, we have two kinds of magnetic fields independent from each other. First the ability of each examinee and second, the difficulty of each problem.

\begin{figure}[ht]
\begin{center}
\includegraphics[width = 8cm]{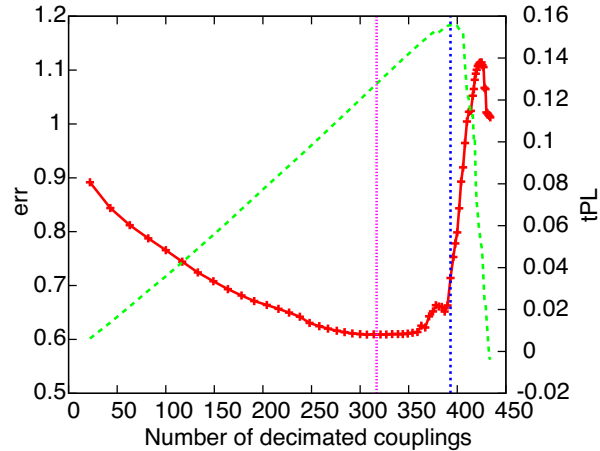}
\caption{PLM+decimation algorithm in the case with $J=500$.
The same symbols and notations are used in Fig. \ref{fig:1}.
The peak of the tilted pseudo likelihood function does not coincide with the minimum point of the error values as depicted by the left vertical line while the right vertical line denotes the maximum point of the tilted pseudo likelihood function in ${\bf w}$.}
\label{fig:15}
\end{center}
\end{figure}

We also notice the following statements on the performance of PLM+decimation.
Our rule to stop the decimation process is to stop it when the maximum of the tilted pseudo likelihood ${\cal PL}_{{\rm tilted}}$ is reached (we call the maximum the terminal point).
However we confirm that, when the number of the given data is small, for instance $J=500$, PLM+decimation algorithm fails to give the better estimation than the optimal case of PLM + $L_1$ estimation.
In this case, the standard use of PLM+decimation algorithm does not lead to the best performance as shown in Fig. \ref{fig:15}.
We observe the discrepancy between the terminal point and the best point, at which the error value in ${\bf w}$ takes its minimum. 
The reason why is the curve of the tilted pseudo likelihood is not sharp as shown in Fig. \ref{fig:15}.
The maximum point of the tilted pseudo function is the farthest location of the likelihood function measured from the line given by $x {\cal PL}_{{\rm max}} + (1-x) {\cal PL}_{{\rm min}}$ for $x \in [0,1]$ to detect the sudden change of its value due to decimation.
When the number of data $J$ is small, the uncertainty of inference remains.
Therefore the pseudo likelihood function does not change drastically depending on decimation of the coefficients.
The uncertainty of estimation thus reflects lack of sharpness in the curve of the pseudo likelihood function.
As a result, we can not detect the best point by detecting the maximum point of the tilted likelihood function.

\begin{figure}[h]
\begin{center}
\includegraphics[width = 14cm]{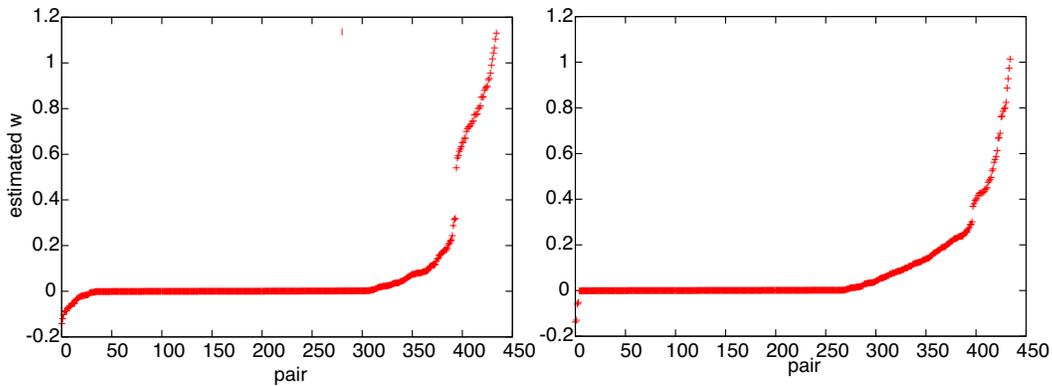}
\caption{The estimation of $w$ in PLM+$L_1$ estimation with $J=1000$ (left panel)
 and $J=500$ (right panel). We plot the sorted data according to the
 estimated value of $w$} 
\label{fig:17}
\end{center}
\end{figure}

Notice that, when we use PLM+$L_1$ estimation to detect the existence of non-zero coefficients in {\bf w}, we must decide a threshold value of the cheating students. 
It is difficult to decide this value without a preliminary knowledge. 
In addition, if the number of tests $J$ is small, this value is more difficult to determine the threshold as shown in Fig. \ref{fig:17}, although we find a clear gap between the several values around zero and the other when the number of tests is a relatively large.

In this sense, both of the method, PLM + decimation and PLM + $L_1$, demands the large number of the data to correctly infer the parameters.
We should emphasize that the PLM + decimation is free from the arbitrariness of the lambda reflecting the performance of the inference and the threshold determining zeros in the parameters. 
As well as its performance as shown in the error between the inferred and predetermined parameters, this advantage point is remarkable efficacy of the decimation algorithm. 
\section{Summary}
We formulated the item response theory with ``cheating students"  in the context of the Boltzmann machine learning. 
We applied the pseudo likelihood estimation to our formulation in order to mitigate the computational complexity to infer the coefficient expressing the degree of cheating on tests and the biases characterizing the difference between the ability of the examinees and the difficulty of the problems which they solve. 
To improve the precision of the estimation and avoid any arbitrariness in the inference, we used PLM+decimation algorithm. 
We contrasted the algorithm with PLM+$L_1$ estimation. 
Both of the approaches are based on sparseness involved in the inference problem. 
We showed that PLM+decimation algorithm, while it does not remain any arbitrariness when we perform it, is comparable or often outperforms PLM+$L_1$ estimation. 
The key point is that tilted pseudo likelihood function is useful to determine when to stop the step of decimation. 
If the number of the training data, namely the number of tests, was small, tilted pseudo likelihood function did not yield the best estimation. 
We hope that the future study finds out a more suitable function than the tilted pseudo likelihood function to decide to terminate decimation steps.
The experiment by use of the actual data is desired to show the performance of our method.

\section*{Acknowledgement}
One of the author M.O. thanks the fruitful discussions with Muneki Yasuda and Kazuyuki Tanaka.
The present work is performed by the financial support from MEXT KAKENHI Grants No. 251200008 and 24740263, and the Kayamori Foundation of Informational Science Advancement.
This work is initiated from homework in the education program ``Student's Innovative Communications and Environment 2013 (SKE2013)" in Kyoto University.
\bibliography{paper_ver5}
\end{document}